\documentclass[twoside,11pt]{article}
\usepackage[preprint,abbrvbib]{jmlr2e}

\usepackage{microtype}
\usepackage[T1]{fontenc}
\usepackage{booktabs}
\usepackage{placeins}
\usepackage{pifont}
\usepackage{xcolor}
\usepackage{listings}

\definecolor{codebg}{HTML}{F8F8F8}
\definecolor{codecomment}{HTML}{6A9955}
\definecolor{codestring}{HTML}{A31515}
\definecolor{codekeyword}{HTML}{0000FF}
\definecolor{codeapi}{HTML}{795E26}
\definecolor{codebuiltin}{HTML}{267F99}
\hypersetup{colorlinks=true, linkcolor={blue!70!black}, citecolor={blue!70!black}, urlcolor={blue!70!black}}

\newcommand{\cmark}{\ding{51}}

\title{xaitimesynth: A Python Package for Evaluating Attribution Methods for Time Series with Synthetic Ground Truth}
\ShortHeadings{xaitimesynth: Evaluating Attribution Methods with Synthetic Ground Truth}{Baer}
\firstpageno{1}

\begin{document}

\author{\name Gregor Baer \email g.baer@tue.nl \\
       \addr Information Systems, Industrial Engineering \& Innovation Sciences\\
       Eindhoven University of Technology \\
       Eindhoven, The Netherlands
       }

\maketitle

\begin{abstract}
Evaluating time series attribution methods is difficult because real-world datasets rarely provide ground truth for which time points drive a prediction.
A common workaround is to generate synthetic data where class-discriminating features are placed at known locations, but each study currently reimplements this from scratch.
We introduce \texttt{xaitimesynth}, a Python package that provides reusable infrastructure for this evaluation approach.
The package generates synthetic time series following an additive model where each sample is a sum of background signal and a localized, class-discriminating feature, with the feature window automatically tracked as a ground truth mask.
A fluent data generation API and YAML configuration format allow flexible and reproducible dataset definitions for both univariate and multivariate time series.
The package also provides standard localization metrics, including AUC-PR, AUC-ROC, Relevance Mass Accuracy, and Relevance Rank Accuracy.
\texttt{xaitimesynth} is open source and available at \url{https://github.com/gregorbaer/xaitimesynth}.
\end{abstract}


\section{Introduction}

Explainable AI (XAI) methods aim to make machine learning predictions interpretable, commonly by producing attributions that indicate how much each input feature contributed to a prediction. As deep learning models see growing adoption for time series classification~\citep{fawaz.etal_2019_deep}, attribution methods are increasingly adapted to temporal data (for a review, see \citet{theissler.etal_2022_explainable}). Yet producing attributions is only half the problem: we also need to evaluate them.

XAI methods are commonly assessed through functional evaluation, where computational metrics quantify desirable properties of explanations~\citep{nauta.etal_2023_anecdotal}. One central property is \emph{correctness}: how accurately an explanation reflects the model's decision-making process. A widely used family of correctness metrics is perturbation-based, measuring how model output changes when supposedly important features are altered. However, these metrics can be sensitive to implementation choices~\citep{schlegel.keim_2023_deep,bluecher.etal_2024_decoupling,simic.etal_2025_comprehensive} and yield systematically different results across classes~\citep{baer.etal_2025_why,baer.etal_2025_classdependent}, raising questions about their reliability as standalone evaluation tools.

Localization-based metrics offer a complementary approach. Instead of perturbing inputs, they measure the spatial agreement between attributions and known ground truth feature locations. Since real-world datasets rarely provide such ground truth, researchers generate synthetic data where class-discriminating features occupy known regions~\citep{ismail.etal_2020_benchmarking,turbe.etal_2023_evaluation,enguehard_2023_learning,serramazza.etal_2024_improving,nguyen.etal_2024_robust,baer.etal_2025_why,nguyen.ifrim_2025_tshap,holzapfel.etal_2025_concept}. This strategy can serve as a sanity check~\citep{adebayo.etal_2018_sanity}.

Synthetic evaluation has been critiqued for not reflecting the model's learned decision boundary~\citep{rawal.etal_2025_evaluating}. A common assumption, also recognized by \citet{nauta.etal_2023_anecdotal}, is that if the only class-discriminating information in the data is the simulated feature and a model achieves high accuracy, it likely relies on that feature. This assumption can break down through shortcut learning: if the generation process introduces unintended artifacts, such as systematic differences in signal statistics between classes, the model may exploit those instead. This is a risk researchers should consider when designing synthetic evaluation setups. Despite these limitations, synthetic data gives researchers full control over the data-generating process, allowing them to investigate attribution behavior in settings where the ground truth is known by construction. In practice, however, each study reimplements synthetic data generation independently.

We introduce \texttt{xaitimesynth}, a Python package that standardizes this evaluation workflow for time series attribution methods. Following the same principle as CLEVR-XAI~\citep{arras.etal_2022_clevrxai}, which provides ground truth feature locations for evaluating visual explanations, the package generates synthetic time series where each sample combines a background signal with a localized, class-discriminating feature, and records the feature location as a ground truth mask. A fluent data generation API and YAML configuration format support reproducible dataset definitions for both univariate and multivariate time series, with fine-grained control over feature placement across channels. The package also provides standard localization metrics, including AUC-PR, AUC-ROC, Relevance Mass Accuracy, and Relevance Rank Accuracy.

\section{Related Work} \label{sec:related}

Several packages produce post-hoc attributions.
Captum~\citep{kokhlikyan.etal_2020_captum} is a widely used library for feature attribution in PyTorch, applicable across data modalities; it also includes perturbation-based evaluation metrics such as infidelity and sensitivity. Since general-purpose explanation libraries typically require adaptation for temporal data, packages like TSInterpret~\citep{hollig.etal_2023_tsinterpret} and Time Interpret~\citep{enguehard_2023_time} extend Captum and other backends to provide unified interfaces for time series attribution methods.

For evaluation, Quantus~\citep{hedstrom.etal_2023_quantus} offers a comprehensive collection of explanation quality metrics, including localization metrics such as those we implement, but targets primarily image data and does not support time series generation. AIX360~\citep{arya.etal_2019_one} includes evaluation metrics such as faithfulness and monotonicity, but not localization-based metrics or synthetic data generation. Among existing packages, Time Interpret~\citep{enguehard_2023_time} comes closest to supporting localization-based evaluation for time series: it provides both perturbation-based and localization metrics alongside its attribution methods. However, it does not include synthetic data generation utilities.

On the data generation side, TimeSynth~\citep{maat.etal_2017_timesynth} generates synthetic time series from configurable signal and noise processes, and GluonTS~\citep{alexandrov.etal_2020_gluonts} includes some synthetic data utilities for forecasting. Neither tracks ground truth feature locations, as they are not designed for XAI evaluation.

In practice, researchers who need localization-based evaluation for time series often implement their own synthetic data pipelines. These implementations vary in complexity and are not always reusable across studies. \texttt{xaitimesynth} fills this gap by combining configurable synthetic time series generation, with automatic ground truth tracking, and standard localization metrics in a single package. It complements existing attribution and evaluation libraries by providing the reusable data generation infrastructure they lack.
Table~\ref{tab:comparison} summarizes the capabilities of the different packages.

\begin{table}[t]
\caption{Comparison of related packages. \texttt{xaitimesynth} combines synthetic time series generation with localization metrics for evaluating attribution correctness.}
\label{tab:comparison}
\centering
\small
\setlength{\tabcolsep}{4pt}
\begin{tabular*}{\textwidth}{@{\extracolsep{\fill}} l c c c c @{}}
\toprule
Package & XAI methods & Perturbation metrics & Localization metrics & Synth.\ time series \\
\midrule
TSInterpret         & \cmark & --- & --- & --- \\
Captum              & \cmark & \cmark & --- & --- \\
AIX360              & \cmark & \cmark & --- & --- \\
Quantus             & --- & \cmark & \cmark & --- \\
Time Interpret      & \cmark & \cmark & \cmark & --- \\
TimeSynth           & --- & --- & --- & \cmark \\
GluonTS             & --- & --- & --- & \cmark \\
\midrule
\texttt{xaitimesynth} & --- & --- & \cmark & \cmark \\
\bottomrule
\end{tabular*}
\end{table}

\section{Package Design} \label{sec:design}

The package has two main components: a data generation module that produces synthetic time series with tracked ground truth masks, and an evaluation module that scores attributions against those masks. We describe each below.

\subsection{Data Generation}
\label{sec:generation}

Following the additive generation approach used in~\citet{baer.etal_2025_why}, each synthetic time series $\mathbf{x} = \mathbf{n} + \mathbf{f}$ is constructed as the sum of a background signal $\mathbf{n}$ and a class-specific feature $\mathbf{f}$, where $\mathbf{n}$ represents a background pattern (e.g., Gaussian noise, a random walk, or a seasonal signal) and $\mathbf{f}$ contains the class-discriminating pattern within a designated time window, with zeros elsewhere.
The package records this window as a binary ground truth mask for every sample.
For multivariate time series, each channel can carry independent signals and features, with optional alignment of feature windows across channels.

\begin{figure}[t]
  \centering
  \includegraphics[width=\linewidth]{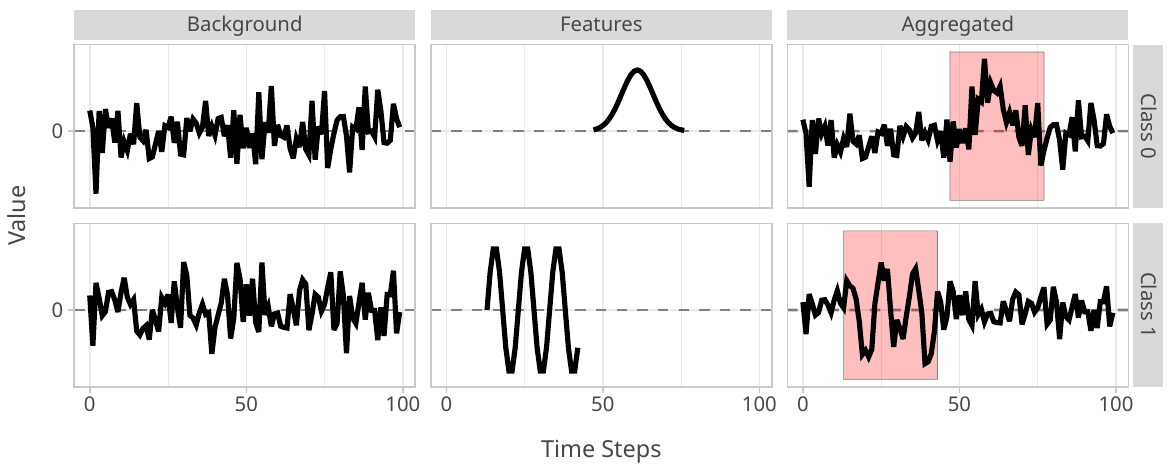}
    \caption{One example per class from the synthetic dataset defined in Listing~\ref{lst:workflow}.
    Each row shows one class; columns show the background signal, the localized feature, and their sum.
    The shaded region in the rightmost column marks the ground-truth feature window.
    Class~0 contains a Gaussian pulse, class~1 an oscillation burst; both share a Gaussian noise background.
    Generated with \texttt{plot\_components(train)}, a built-in visualization utility.}
  \label{fig:example}
\end{figure}

The builder API lets the user declaratively specify signals and features for each class; the package handles ground truth tracking, normalization, and output formatting internally.
The following example in Listing~\ref{lst:workflow} defines a two-class dataset, generates train and test splits, and evaluates attributions against the ground truth masks. Figure~\ref{fig:example} shows one example per class from this dataset.

\begin{lstlisting}[float=t, caption={Example workflow: defining a two-class synthetic dataset, generating train/test splits, and evaluating attributions against ground truth masks.}, label=lst:workflow]
from xaitimesynth import TimeSeriesBuilder, gaussian_noise, gaussian_pulse, seasonal
from xaitimesynth.metrics import auc_pr_score, relevance_mass_accuracy

# Define dataset settings
base_builder = (
    TimeSeriesBuilder(n_timesteps=100, normalization="zscore")
    .for_class(0)
    .add_signal(gaussian_noise(sigma=1.0))
    .add_feature(
        gaussian_pulse(amplitude=3.0),
        random_location=True,
        length_pct=0.3,
    )
    .for_class(1)
    .add_signal(gaussian_noise(sigma=1.0))
    .add_feature(
        seasonal(period=10, amplitude=3.0),
        random_location=True,
        length_pct=0.3,
    )
)

# Generate datasets with different seeds and sample counts
train = base_builder.clone(n_samples=200, random_state=42).build()
test = base_builder.clone(n_samples=50, random_state=43).build()

# Obtain attributions from your XAI method; 
# (shape: (n_samples, n_dims, n_timesteps)
attributions = xai_method.explain(test["X"]) # pseudo-code

# Evaluate against ground truth
auc = auc_pr_score(attributions, test, normalize=True)
rma = relevance_mass_accuracy(attributions, test)
\end{lstlisting}

Built-in signal types include Gaussian noise, uniform noise, red noise, random walk, seasonal, and trend; built-in feature types include peak, trough, and Gaussian pulse.
While the package distinguishes between signals and features, all generators can be used in either role, so researchers can flexibly compose the data-generating process to match their experimental needs.
A \texttt{manual} component accepts arbitrary generator functions, and custom components can be registered via a decorator API, making the package extensible without modifying source files.
Dataset definitions can alternatively be written in YAML and loaded with \texttt{load\_builders\_from\_config()}, which supports YAML anchors for sharing class configurations across datasets.

\subsection{Evaluation Metrics}
\label{sec:metrics}

The included metrics measure the overlap between attribution scores and the binary ground truth masks from the build step, quantifying whether high attributions concentrate in the correct regions.
Table~\ref{tab:metrics} lists the included metrics.

\begin{table}[t]
\caption{Localization metrics currently included in \texttt{xaitimesynth}.}
\label{tab:metrics}
\centering
\small
\setlength{\tabcolsep}{4pt}
\begin{tabular*}{\textwidth}{@{\extracolsep{\fill}} l p{3.8cm} l @{}}
\toprule
Metric & Function & Reference \\
\midrule
AUC-ROC                          & \texttt{auc\_roc\_score}            & \citet{fawcett_2006_introduction} \\
AUC-PR                           & \texttt{auc\_pr\_score}             & e.g., \citet{baer.etal_2025_why} \\
Relevance Mass Accuracy          & \texttt{relevance\_mass\_accuracy}  & \citet{arras.etal_2022_clevrxai} \\
Relevance Rank Accuracy          & \texttt{relevance\_rank\_accuracy}  & \citet{arras.etal_2022_clevrxai} \\
Pointing Game                    & \texttt{pointing\_game}             & \citet{zhang.etal_2018_topdown} \\
Normalized Attribution Correspondence    & \texttt{nac\_score}                 & \citet{peters.etal_2005_components} \\
MAE                              & \texttt{mean\_absolute\_error}      & --- \\
MSE                              & \texttt{mean\_squared\_error}       & --- \\
\bottomrule
\end{tabular*}
\end{table}

AUC-ROC and AUC-PR treat attributions as scores and the ground truth mask as labels, measuring ranking quality; both support normalization to remove the effect of ground truth prevalence.
Relevance Mass Accuracy and Relevance Rank Accuracy measure overlap between the highest-attributed timesteps and the ground truth window at different granularities: the former computes the fraction of total attribution mass inside the ground truth, while the latter checks whether the top-$K$ ranked timesteps fall within it.
The Pointing Game reduces this to a single binary check on the maximum attribution.
Normalized Attribution Correspondence, adapted from the normalized scanpath saliency (NSS) metric in eye-tracking, measures the mean z-scored attribution at ground truth locations.
MAE and MSE treat evaluation as regression, measuring pointwise error between attributions and the binary mask.

\section{Availability and Documentation}
\label{sec:availability}

\texttt{xaitimesynth} is released under the MIT license and available at \url{https://github.com/gregorbaer/xaitimesynth}. It can be installed via pip:

\begin{lstlisting}[language=bash]
pip install xaitimesynth
\end{lstlisting}

The package requires Python 3.10 or later.
Runtime dependencies are NumPy, pandas, PyYAML, and lets-plot; no deep learning framework is required for data generation or metric computation.
Documentation, including usage guides, API reference, and YAML configuration examples, is available at \url{https://gregorbaer.github.io/xaitimesynth/}.
The documentation also covers output data shapes and worked examples for both univariate and multivariate settings.
A persistent archive is available at \url{https://doi.org/10.5281/zenodo.18888778}.

\section{Conclusion}
\label{sec:conclusion}

\texttt{xaitimesynth} standardizes localization-based evaluation of time series attribution methods by combining synthetic data generation with automatic ground truth tracking and standard localization metrics, replacing per-study reimplementation with a shared toolkit.
The builder API and YAML configuration format make dataset definitions concise, reproducible, and straightforward to share alongside code.

\acks{This work is supported by the European Union's HORIZON Research and Innovation Program under grant agreement No.\ 101120657, project ENFIELD (European Lighthouse to Manifest Trustworthy and Green AI).}

\bibliography{references.bib}

\end{document}